%% file: main.tex
\documentclass[10pt,twocolumn,letterpaper]{article}

\usepackage[pagenumbers]{cvpr} 

\usepackage{graphicx}
\usepackage{amsmath}
\usepackage{amssymb}
\usepackage{booktabs}

\usepackage{multirow}
\usepackage{graphicx}
\usepackage{arydshln}
\usepackage[normalem]{ulem}
\useunder{\uline}{\ul}{}

\usepackage[pagebackref,breaklinks,colorlinks]{hyperref}

\usepackage[capitalize]{cleveref}
\crefname{section}{Sec.}{Secs.}
\Crefname{section}{Section}{Sections}
\Crefname{table}{Table}{Tables}
\crefname{table}{Tab.}{Tabs.}

\def\cvprPaperID{8119} 
\def\confName{CVPR}
\def\confYear{2023}

\usepackage{float}

\begin{document}

\title{ObjCAViT: Improving Monocular Depth Estimation Using Natural Language Models And Image-Object Cross-Attention}

\author{Dylan Auty$^1$ and Krystian Mikolajczyk$^2$\\
Imperial College London\\
Exhibition Rd, South Kensington, London SW7 2BX\\
{\tt\small \{dylan.auty12$^1$, k.mikolajczyk$^2$\}@imperial.ac.uk}
}
\maketitle

\input{sections/abstract}


\input{sections/introduction}
\input{sections/background}

\input{sections/method}

\input{sections/experiments}
\input{sections/conclusion}

{\small
\bibliographystyle{ieee_fullname}
\bibliography{references}
}

\appendix
\input{sections/app-positional-encoding-arch}
\input{sections/app-additional-examples.tex}

\end{document}

%% file: sections/abstract.tex
\begin{abstract}
    While monocular depth estimation (MDE) is an important problem in computer vision, it is difficult due to the ambiguity that results from the compression of a 3D scene into only 2 dimensions. It is common practice in the field to treat it as simple image-to-image translation, without consideration for the semantics of the scene and the objects within it. In contrast, humans and animals have been shown to use higher-level information to solve MDE: prior knowledge of the nature of the objects in the scene, their positions and likely configurations relative to one another, and their apparent sizes have all been shown to help resolve this ambiguity.

    In this paper, we present a novel method to enhance MDE performance by encouraging use of known-useful information about the semantics of objects and inter-object relationships within a scene. Our novel ObjCAViT module sources world-knowledge from language models and learns inter-object relationships in the context of the MDE problem using transformer attention, incorporating apparent size information.
    Our method produces highly accurate depth maps, and we obtain competitive results on the NYUv2 and KITTI datasets. Our ablation experiments show that the use of language and cross-attention within the ObjCAViT module increases performance. Code is released at \url{https://github.com/DylanAuty/ObjCAViT}.

\end{abstract}

%% file: sections/introduction.tex
\section{Introduction}
\label{sec:introduction}
\begin{figure}[h]
    \centering
    \includegraphics[width=\linewidth]{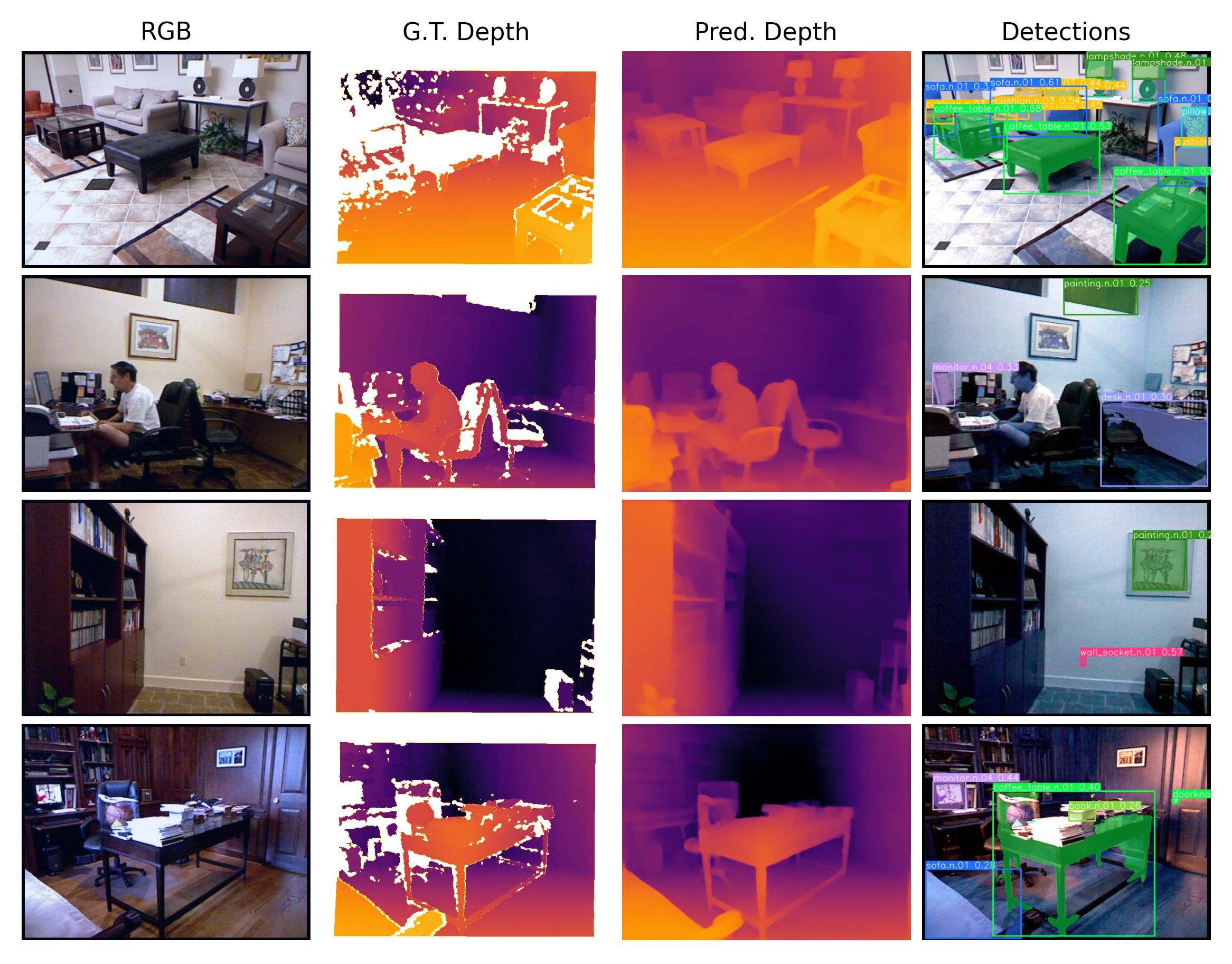}
    \caption{Example inputs, ground-truth depth maps, and predicted depth outputs from our best model running on the NYUv2 dataset, including the YOLOv7 detections that we use as auxiliary information to improve performance. By using object semantics and modelling the relationships \textit{between} objects, our model is able to improve depth estimation performance.}
    \label{fig:eg_output}
\end{figure}

Monocular depth estimation (MDE) is an important problem in the field of computer vision, but is fundamentally ill-posed: a 2D image may have been taken from a number of 3D scenes. Because of this ambiguity, all MDE systems must inherently make assumptions about the world, aiming to estimate the most likely depth values, given the set of assumptions that they hold, from the input image. In the biological domain, many of these assumptions about the input and the world have been identified, and are known as depth ``cues". These cues range from the obvious, for example the effect of perspective on the apparent size of an object, to the more implicit or reasoning-based, such as the inference of the likely true size of an unknown object based on the semantic relationship it appears to have with the objects in its vicinity.

Many works in the literature have tackled the problem of MDE with good results, but almost all assume that the information required to make assumptions about scene geometry will be learned implicitly. Because the poorly-posed nature of the problem demands that assumptions of some description be made to resolve ambiguity, the implication is that any MDE system must be expending effort on working out an appropriate set of assumptions before it can begin to detect those assumptions in an input image and learn appropriate mappings from them to the likely output depth values. This presents an obvious inefficiency, wasting limited model capacity and training data on learning the \textit{existence} of relevant assumptions, rather than how to \textit{exploit} them for the end problem. Being able to incorporate prior information that is known already to be useful for MDE to an MDE model would therefore be expected to produce a boost in performance as the model no longer has to expend capacity learning these cues, provided that the assumptions that the model is guided towards by design are at least as useful as those it may learn on its own.

The mechanism of cross-attention has been used to great effect to enhance performance in other tasks, but has not (to the best of our knowledge) been exploited to introduce non-image features for MDE. It provides a means to modify one set of features based on another set of features, for instance fine-grained image features based on coarse-grained image features \cite{chen_crossvit_2021}, or images of text with actual text \cite{li_selfdoc_2021}. In the motivation for our work, we consider the role that inter-object relationships may play in improving the way in which image features are represented for the task of MDE.

Modifying image features based on object and inter-object-relationship semantics fits perfectly with the purpose and intuition behind cross-attention. The ambiguity of the MDE problem requires layers of interpretation to resolve: first interpreting form and semantic meaning from objects in a scene, then comparing them to previous understanding of their likely shapes, sizes, and co-occurrences, and finally relating them to the image in which they are observed to give further context on their likely position in the scene. This interpretation of the semantics and the relationships between objects, their known sizes, their apparent sizes, and relationships to each other, has been shown to be present in animal vision systems \cite{sousa_judging_2011, hershenson_pictorial_1998} and, to an extent, in machine vision systems \cite{auty_monocular_2022}. Our work makes use of the cross-attention mechanism to allow the model to learn these known-useful relationships.

In this work, we introduce a novel method to \textit{explicitly} leverage known-useful information about the world for the problem of MDE. In particular, we introduce a language and transformer attention mechanism by which the model is encouraged to focus on not only the visual features of an image, but the semantics of the objects contained within it.

Our main contributions are:
\begin{enumerate}
    \item We introduce a novel method of incorporating auxiliary object detection information to an MDE pipeline with our \textbf{Object Cross-Attention Vision Transformer (ObjCAViT block)}, which allows the model to learn about both the semantic roles of an object in the world and the relationships between those objects that appear in an input scene,
    \item We demonstrate the novel use of joint size-and-position embeddings for object and image patch self-attention with transformers, showing an improvement in depth inference by making use of apparent object sizes,
    \item Our method is able to exploit auxiliary information from frozen off-the-shelf language models and object detectors to improve depth estimation,
    \item We demonstrate clear improvement in performance over the baseline on both the indoor and outdoor domains using the NYUv2 and KITTI datasets respectively.
\end{enumerate}

%% file: sections/background.tex
\section{Background}
\label{sec:background}
\begin{figure*}[h]
    \centering
    \includegraphics[width=\textwidth]{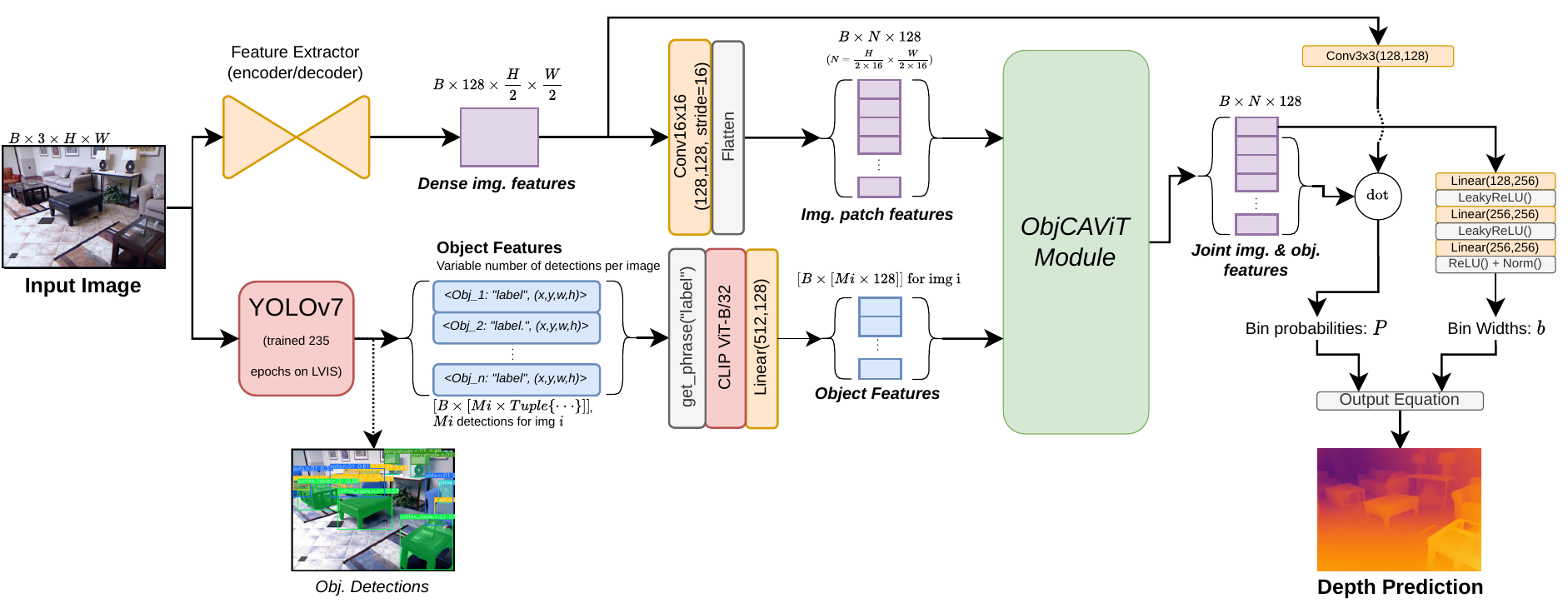}
    \caption{\textbf{An overview of our architecture}, showing the input RGB image on the left and the predicted depth map on the right. The ObjCAViT block is shown in more detail in figure \ref{fig:objcavit-detail}. Orange blocks are learnable, red are pretrained and frozen, and grey are non-learned functions. Image and object features are coloured purple and blue respectively.}
    \label{fig:main_arch}
\end{figure*}

\textbf{Monocular Depth Estimation} (MDE) methods generally treat the problem as a dense image to image mapping. They often consist of a convolutional encoder and decoder, and frequently make use of encoders that have been pretrained on ImageNet \cite{deng_imagenet:_2009} such as ResNet \cite{he_deep_2016}, VGG \cite{simonyan_very_2015}, or EfficientNet \cite{tan_efficientnet_2019}. Once dense features have been extracted from these encoders, they are then upsampled or decoded to provide an output depth map.

Previous SOTA methods have expanded on this basic template in different ways. \cite{eigen_depth_2014, eigen_predicting_2015, lee_big_2019} use a multi-scale prediction approach. \cite{lee_big_2019} make the assumption that the world is (locally) planar. Some \cite{ramamonjisoa_sharpnet:_2019, jiao_look_2018, bai_monocular_2019} co-predict depth with other tasks such as semantic segmentation or surface normals, sharing some weights between the different decoder heads or enforcing consistency between tasks in the loss function.

More recent methods have reframed the problem as classification, where the model must assign each pixel of the input image to a depth ``bin" instead of giving a scalar value. \cite{fu_deep_2018} directly assign each pixel to a bin, giving good performance but producing artifacts due to the discrete nature of the output. AdaBins \cite{bhat_adabins_2020} expanded on this work significantly: rather than using uniform bins, a transformer \cite{vaswani_attention_2017} model was used to adaptively change the distribution and width of the bins according to the input image, assigning final depth values by smoothly interpolating between bin centres using a pixel's bin probability vector. \cite{li_binsformer_2022} improve on adaptive binning by using intermediate features from the image encoder to assist in the binning process, and by adding an auxiliary scene classification task to encourage the model to learn global semantic information. \cite{li_depthformer_2022} use both a transformer and a traditional convolutional encoder to extract image features, before merging them and decoding to a depth map using a transformer decoder. Very recently, \cite{agarwal_attention_2022} have used an all-transformer architecture to extract image features and to perform adaptive binning based on these features. All of these recent methods, however, do not explicitly steer the model towards the learning of local scene or object \textit{semantics}.

\textbf{Cross attention} as a means of combining features from multiple sources is a technique that has produced positive results in a variety of problems, beginning with the original transformer paper \cite{vaswani_attention_2017}. \cite{li_selfdoc_2021} apply cross-attention to a joint visual and textual interpretation of a document. \cite{chen_crossvit_2021} use it for the combination of different scales of image patches in a vision-transformer-like model. In the problem of local feature matching, \cite{sarlin_superglue_2020} applied cross-attention to cause the model to learn the correspondences between pairs of features. While \cite{li_depthformer_2022} make use of deformable cross-attention, it is used to combine two different sets of image features together rather than to introduce non-image information.

\noindent\textbf{Language and depth.} Language and depth is a relatively new area of research. Intuitively, a language model exists to encode human language; if human language in turn encodes human ideas, then it would follow that language models encode human biases. \cite{henlein_what_2022} showed that BERT \cite{devlin_bert_2019} embeddings contain information about the likely distribution of objects across different rooms, the relationship between parts of objects and the objects they are likely part of, as well as the likely target objects for a given verb. \cite{auty_monocular_2022} make use of GloVe \cite{pennington_glove_2014} embeddings as a means of introducing knowledge about the world to the model to encourage it to use the familiar size biological depth cue. CLIP \cite{radford_learning_2021} is a transformer-based model that, given enough data, is able to correlate high-level visual concepts with language; this has been used for image generation from language \cite{ramesh_hierarchical_2022}, and has more recently been applied to zero-shot depth estimation \cite{zhang_can_2022} by correlating image features with the features of depth-related sentences. Despite the difficulty of training CLIP, pre-trained checkpoints for general purpose models are available online. For these reasons, coupled with CLIP's demonstrated high performance on tasks (notably image synthesis) that require a significant degree of world-knowledge, our method uses a frozen, pretrained CLIP model.

%% file: sections/method.tex
\section{Method}
\label{sec:method}


In this section, we describe our method in detail. We begin by motivating the approach taken, then describe the architecture and novel ObjCAViT block used.

The primary motivation of our method is the known relevance, at least in animal vision systems, of \textbf{object semantics} and \textbf{inter-object relations} to the problem of monocular depth estimation (MDE) as clues for depth prediction disambiguation. Without sufficient information to create a one-to-one mapping between observations and 3D scene geometry,
the input image's contents must be interpreted semantically by the model, in the context of its existing assumptions about the nature of the world, to disambiguate possible depth values.
While there are many different monocular depth cues that have been shown to improve performance in animals, we focus here on the semantics and assumptions that may be made about both (a) objects and (b) the relationships between objects in a scene.

While existing methods make the assumption that their model will implicitly learn the nature of any depth cues present in the target domain, we hypothesise that this is a source of \textbf{inefficiency} in a model: expending the limited resources of model capacity and training data on implicit learning of the \textit{existence} of depth cues means that those resources are not being spent on \textit{interpreting} those depth cues. Models that learn to perform MDE implicitly must first infer that perspective (for instance) exists at all, and only then can they begin learning the functions required to map the perspective-related features they observe to depth predictions. Thus, our method is designed around encouraging a model to focus on learning these known-useful relationships; provided the bias deliberately introduced is at least as useful as that that would be learned by the model implicitly, an improvement in performance would be expected.

Following this motivation, we divide our main model into different sections: the \textbf{visual encoder}, the \textbf{object encoder}, and the \textbf{ObjCAViT block}.

\subsection{Visual Encoder}
\label{sec:visual-encoder}
    The visual encoder is responsible for interpreting the raw RGB pixel values from the input image. In our architecture, we use a convolutional encoder-decoder model to extract dense image features. The architecture of this encoder-decoder mirrors that of AdaBins \cite{bhat_adabins_2020}: it comprises an EfficientNet-B5\cite{tan_efficientnet_2019} encoder pretrained on ImageNet \cite{deng_imagenet:_2009}, and a standard upsampling decoder with skip connections from the encoder. The visual encoder outputs features at $0.5\times$ the resolution of the input image. In the style of the vision transformer \cite{dosovitskiy_image_2021}, features are then divided into 16x16 patches and further embedded using a 2d convolution with kernel size and stride both set to 16x16, and the resulting image patch features are flattened for use by the ObjCAViT block.
    
\subsection{Object Encoder}
\label{sec:object-encoder}
    The object encoder is responsible for both extracting and then providing embeddings for the objects in the scene.
    
    \textbf{Object detection.} To simplify the task of the depth estimation network by avoiding having to learn multiple tasks, we use a frozen YOLOv7 object detector\cite{wang_yolov7_2022} to extract object bounding boxes and class labels. In our implementation, we use the segmentation-capable model, first pretrained on the 80-class MS-COCO \cite{lin_microsoft_2015} dataset, and then fine-tuned for 235 epochs on the 1203-class LVIS \cite{gupta_lvis_2019} v1.0 instance and semantic segmentation dataset.
    LVIS is used due to the large number of labels available, and because labels are given as Wordnet \cite{miller_wordnet_1995} synsets to remove the problem of polysemy.
    
    \textbf{Language embedding.} The purpose of introducing object information is to make use of a pre-existing model of the underlying nature of that object (e.g. its shape, size, or common co-occurrences with other objects). In our method, this prior world-knowledge is extracted from the language model used in CLIP \cite{radford_learning_2021}. Existing work has shown that CLIP, as well as other language models, encode likely spatial and depth information about an object (see section \ref{sec:background}), and thus we use CLIP language embeddings as a proxy for general world knowledge.
    
    WordNet definitions are extracted for the class label of each detected object\footnote{With one exception, ``stop\_sign.n.01", the definition for which is sourced from the first sentence of the English language Wikipedia article for "stop sign".} in the image. These definitions are then used as the language input to a frozen pretrained CLIP ViT-B/32 language encoder, which outputs a single $512d$ embedding. These are then reduced with a linear layer to match the dimensionality of the image features.
    
    We experiment with two different language templates for a given object label. The first, \textit{def. only}, uses only the WordNet definition for that object's synset label $\{Obj_i\}$ as the language input. The second, \textit{def.+sz\_rel}, aims to extract information about object relationships from the language model using a relative size clause, which selects one other random object $\{Obj_j\}$ detected in the scene, compares its apparent size (bounding box area) to that of the current object $Obj_i$, and then compares them on a 7-point scale of size comparison phrases: 
    \textit{[much smaller/smaller/a bit smaller than, about the same size as, a bit bigger/bigger/much bigger than]}
    The final template has the form \textit{``This is a/an $\{Obj_i\}$, defined as \{definition\}. This $\{Obj_i\}$ appears to be \{size comparison\} the \{$Obj_j$\}."}. The comparative scale text is chosen according to the bounding box area factor $f_A = \frac{A_i}{A_j}$ for the areas of objects $i$ and $j$. We assign $f_A$ to bins: two edge bins for $f_A \le e^{-3}$ and $f_A > e^3$, and 5 bins with centres defined at $e^k \forall k \in \{-2, -1, 0, 1, 2\}$. These 7 bins are each assigned a natural language phrase from the 7-point scale accordingly, to give the final phrase.
    
\subsection{ObjCAViT Module}
\label{sec:objcavit-module}
    \begin{figure}
        \centering
        \includegraphics[width=\linewidth]{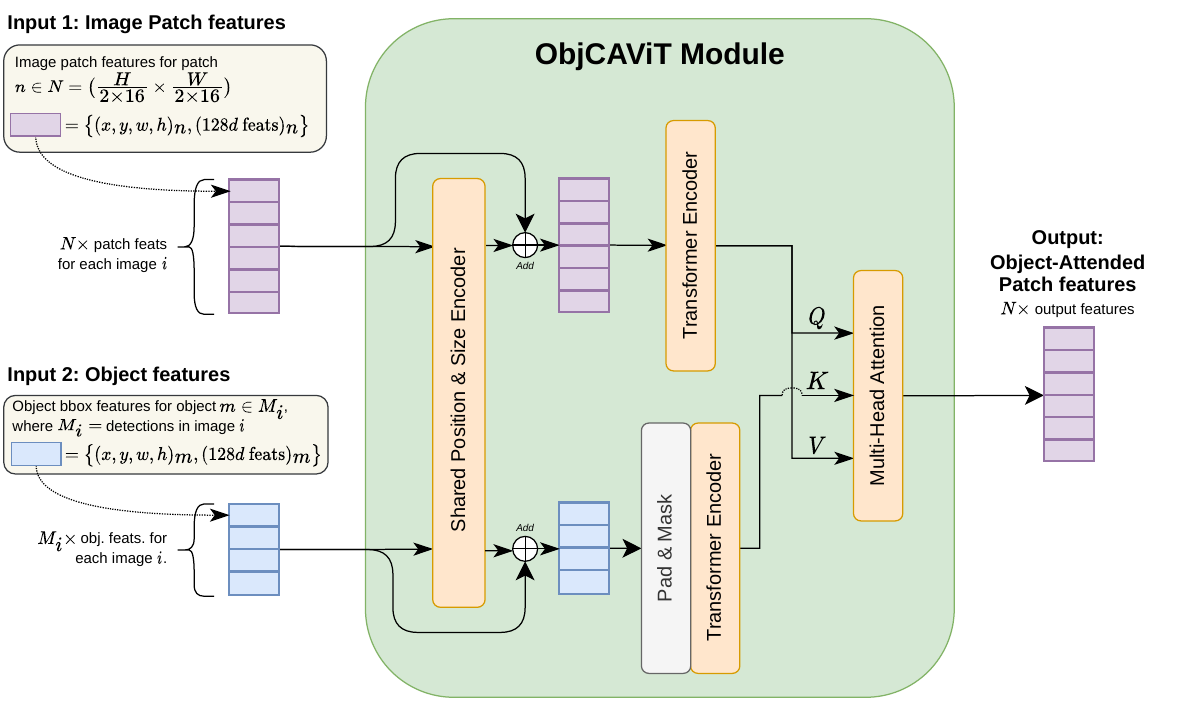}
        \caption{\textbf{Our proposed ObjCAViT block}, which uses self-attention to allow the model to learn to relate image features to each other and to learn the relationships between detected objects in a scene. It then combines the two using a cross-attention mechanism that modifies the image features based on the learned relationships between the detected objects.
        }
        \label{fig:objcavit-detail}
    \end{figure}
    
    \begin{figure*}[!h]
        \centering
        \includegraphics[width=\textwidth]{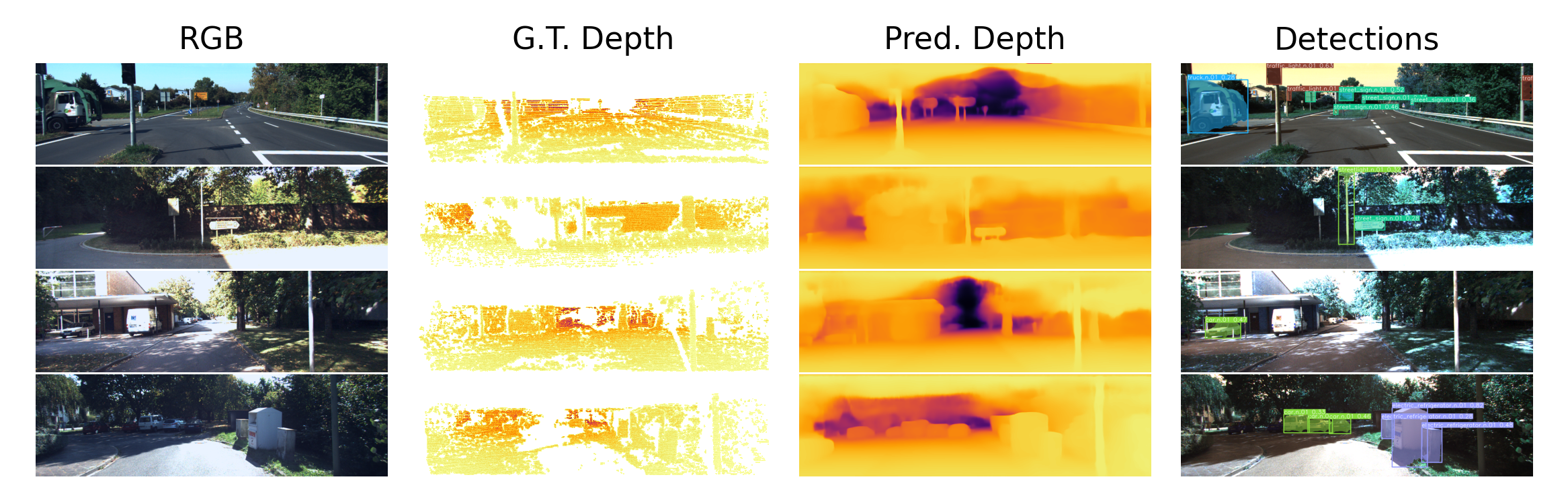}
        \caption{Example inputs and outputs of our best model, demonstrated on the KITTI dataset. This model uses only synset definitions for the detected objects, and uses both object bounding box (or image patch) centre coordinates and dimensions as inputs to the positional encoders when performing self-attention. The object detections output by YOLOv7 are also shown in the right-hand column.}
        \label{fig:pred1-kitti}
    \end{figure*}
    
    The \textbf{Object Cross-Attention Vision Transformer}, or \textbf{ObjCAViT}, is a principal contribution of our work, and its architecture can be seen in figure \ref{fig:objcavit-detail}. It is an extension of the Vision Transformer \cite{dosovitskiy_image_2021} and miniViT \cite{bhat_adabins_2020} of previous work, but with the key difference that it allows the model to learn three main cues: the relationships between objects, the relationships between patches of the input image, and how to link the two together. This mirrors the MDE cues known to be present in the biological domain (see section \ref{sec:background}).

    First, the image and object features are fed through the common positional encoder, and the resultant positional embeddings are added to the image/object features. We tried different positional encoders: (1) a simple MLP with LeakyReLU activation that takes the $(x,y)$ coordinates and outputs a $128d$ positional embedding, (2) a similar model but with 4 inputs instead of 2 to allow embedding of bounding box/patch dimensions as well as positions.
    
    The object features are self-attended to using a transformer encoder with 4 layers of 4 heads each, a feedforward dimension of 1024, and an embedding dimension of 128. The visual patch features are similarly self-attended. This yields a sequence of features representing the visual aspects of the image, and another set of features representing the semantics of the objects in the image and their relationships to one another. Lastly, a cross-attention mechanism is used, with the visual features acting as the query and value inputs, and the object features acting as the key input. This allows the network to attend to the image features based on the object features. The ObjCAViT module outputs a sequence of features.
    
\subsection{Output Stage}
\label{sec:output-stage}
    The first feature in the ObjCAViT output sequence is processed by a simple MLP to obtain a vector of 256 depth bin widths, $b$, for that input image. The next 128 features in the sequence are dotted with the dense image features to give bin probabilities $P$ for each pixel of the dense image feature map. This output format, and the post-processing we use to convert it into depth maps, follow \cite{bhat_adabins_2020, li_binsformer_2022}.
    
    To obtain the depth, the bin centres $c$ are computed from the bin widths $b$ and the maximum and minimum depth values $d_{max}$ and $d_{min}$ in use for the dataset:
    \begin{equation}
        c(b_i) = d_{min} + (d_{max} - d_{min})\biggl(\frac{b_i}{2} + \sum_{j=1}^{i-1}b_j\biggr)
    \end{equation}
    The final depth value $d$ for a pixel $n$ of the dense feature map $P$ is then obtained by summing the bin centres, weighted by the bin probabilities across the 256 bins:
    \begin{equation}
        d_{n} = \sum_{i=1}^{256}c(b_i)P_n
    \end{equation}
    
    Because $P$ is at $0.5 \times$ input resolution, the final output depth map is as well. We bilinearly upsample this output to match the input image resolution for loss computation.

\subsection{Loss Function}
\label{sec:loss-function}
    The loss functions used are the scale-invariant log loss and the bin centre density loss of AdaBins \cite{bhat_adabins_2020}. The ground-truth and predicted depth values for pixel $n$ are given as $d^*_n$ and $d_n$ respectively. The depth loss we use is given by:
    \begin{equation}
        \label{eq:depth-loss}
        \mathcal{L}_{SILog} = 10 \sqrt{\frac{1}{N}\displaystyle\sum_{i \in N} g_{n}^2 + \frac{0.15}{N^2}(\displaystyle\sum_{n \in N} g_{n})^2}
    \end{equation}
    where $g_{n} = log(d_{n}) - log(d^*_{n})$ and $N$ is the total number of pixels with valid depth values. The bin centre density loss encourages the network to predict bin centres that are similarly distributed to the ground-truth depth values. It is given by the Chamfer loss between the set of depth values in the ground truth $d$ and the bin centres $c(b)$, and vice versa:
    \begin{equation}
        \label{eq:bin-centre-loss}
        \mathcal{L}_{bin} = \text{chamfer}(d,c(b)) + \text{chamfer}(c(b),d)
    \end{equation}

%% file: sections/experiments.tex
\section{Experiments}
\label{sec:experiments}
In this section, we detail our experiments. In section \ref{sec:setup}, we describe our baseline, datasets, hyperparameters and training procedures, and evaluation metrics. In section \ref{sec:results}, we present our results, and include ablation studies to demonstrate the effectiveness of our use of language embeddings, our use of self-attention between object features to allow the model to learn the relationships between the different objects detected in the scene, and finally to prove the efficacy of our positional embedding strategy. We show qualitative results on KITTI from our best model in figure \ref{fig:pred1-kitti}, and from NYUv2 in figure \ref{fig:eg_output}.

\subsection{Experimental Setup}
    \label{sec:setup}
    \subsubsection{Baseline}
    \label{sec:baseline}
        Our baseline architecture mirrors that of AdaBins, with identical image extractors and treatment of image patch features, but without any object detections, language embeddings, object features/self-attention, or cross-attention between image patch and object features. We report baseline results using our own reimplementation of AdaBins for consistency. Our method may be considered an extension of AdaBins, and our experiments will show the effect of progressively adding each part of our method.

    \subsubsection{Datasets}
    \label{sec:datasets}
        Our experiments are performed on two datasets, to show applicability across domains: NYUv2 \cite{silberman_indoor_2012} for indoor and KITTI \cite{geiger_vision_2013} for outdoor. We use the same split of these datasets as used in \cite{bhat_adabins_2020}, and we use the same training and testing crops and augmentations as that work also. Our model outputs images at half-resolution relative to the input image dimensions; we therefore perform bilinear upsampling back to the input resolution as a final step after output.
    
        \textbf{NYUv2} is an indoor dataset across a variety of different indoor scenes, with (mostly) dense depth ground-truth captured with a Microsoft Kinect. We use 24231 training examples and the official test split of 654 examples. We train on random crops of size $[416\times544]$, and evaluate on the Eigen crop \cite{eigen_depth_2014}. The input images at test time have dimensions $[480\times640]$. When evaluating, we follow convention \cite{bhat_adabins_2020, lee_big_2019}: we take predictions from each input image and its mirror. We then re-mirror the mirror prediction and average it with the non-mirrored prediction to give the final result.
            
        \textbf{KITTI} is an outdoor driving dataset with sparse depth ground-truth captured using LiDAR. We use 23157 training and 696 test examples. Following \cite{bhat_adabins_2020}, during training we take a random crop of size $[352\times704]$, and during evaluation we follow the cropping strategy from \cite{garg_unsupervised_2016}. We perform the same mirror-image prediction and averaging for KITTI as we do for NYUv2.
        
    \subsubsection{Training Procedure}
        We implement all our experiments using PyTorch. All results are reported after 25 epochs of training on two GPUs with 24Gb+ of VRAM (either 2x Quadro RTX6000s or 1x RTX A5000 + 1x RTX A6000, all by NVIDIA), using a batch size of 8. Following the procedure detailed in \cite{bhat_adabins_2020}, we use the AdamW optimizer with a learning rate of 0.000357 and a OneCycleLR learning rate scheduler, with max\_lr=0.000357, cycle\_momentum=True, base\_momentum=0.85, max\_momentum=0.95, div\_factor=25, and final\_div\_factor=100.
        
    \subsubsection{Evaluation Metrics}
    \label{sec:metrics}
        The metrics used for evaluation are those defined by \cite{eigen_depth_2014}: Abs relative difference (Abs or Abs Rel): $\frac{1}{T}\sum_{i=1}^{T} \frac{|d_i - d_i^*|}{d_i^*}$, Squared relative difference (Sq or Sq Rel): $\frac{1}{T}\sum_{i=1}^{T} \frac{\|d_i - d_i^*\|}{d_i^*}$, lin. RMSE (RMS): $\sqrt{\frac{1}{T}\sum_{i=0}^{T}\|d_i - d_i^*\|^2}$, log RMSE (RMSL): $\sqrt{\frac{1}{T}\sum_{i=0}^{T}\|log(d_i) - log(d_i^*)\|^2}$, and the threshold accuracy $\delta_n$: \(\%\) of \(d_i\) s.t. $max(\frac{d_i}{d_i^*}, \frac{d_i^*}{d_i}) = \delta < thr$, where $\delta_n$ denotes that $thr = 1.25^n$ (we use $n \in \{1, 2, 3\}$).
    
        We follow the accepted convention in other works \cite{bhat_adabins_2020, lee_big_2019} of computing a running average of each of these metrics. We run inference on a single GPU one example at a time from the evaluation split. Each metric has a running average accumulator $ave_n$, where $n$ is the number of examples seen thus far. We run inference for image $n$ and compute a metric $val_n$. We then update $ave$ according to $ave_{n+1} = \frac{(ave_n \times n) + val_n }{n+1}$.
        
\subsection{Results}
\label{sec:results}
    \begin{table*}[h!]
            \centering
            \resizebox{0.925\textwidth}{!}{%
            \begin{tabular}{c|ll|ccccc|ccc}
            \multicolumn{1}{l|}{Dataset} & Model & Pos. embedding & $\downarrow$ Abs. Rel & $\downarrow$ Sq. Rel & $\downarrow$ RMS & $\downarrow$ RMSL & $\downarrow$ Log10 & $\delta ^1$ & $\delta ^2$ & $\delta ^3$ \\ \hline
            \multirow{4}{*}{\textbf{NYUv2}} & \textit{Baseline (AdaBins reimpl.)} &  & 0.123 & 0.077 & 0.428 & 0.028 & 0.052 & 0.858 & 0.978 & 0.995 \\
             & \textit{Ours (ctrl. zeros)} & pos+bbox\_wh & 0.104 & 0.058 & \textbf{0.367} & \textbf{0.021} & \textbf{0.044} & 0.898 & \textbf{0.985} & \textbf{0.997} \\
             & \textit{Ours (def. $+$ sz\_rel)} & pos+bbox\_wh & 0.104 & 0.059 & 0.373 & 0.022 & \textbf{0.044} & 0.898 & 0.984 & \textbf{0.997} \\
             & \textit{Ours (def. only)} & pos+bbox\_wh & \textbf{0.103} & \textbf{0.056} & 0.368 & \textbf{0.021} & \textbf{0.044} & \textbf{0.902} & \textbf{0.985} & \textbf{0.997} \\ \hline
            \multirow{4}{*}{\textbf{KITTI}} & \textit{Baseline (AdaBins reimpl.)} &  & \textbf{0.058} & 0.192 & 2.355 & 0.009 & \textbf{0.025} & 0.965 & 0.995 & \textbf{0.999} \\
             & \textit{Ours (ctrl. zeros)} & pos+bbox\_wh & 0.059 & 0.200 & 2.397 & 0.009 & 0.026 & 0.962 & 0.994 & \textbf{0.999} \\
              & \textit{Ours (def. $+$ sz\_rel)} & pos+bbox\_wh & \textbf{0.058} & 0.188 & 2.304 & 0.009 & \textbf{0.025} & 0.965 & 0.995 & \textbf{0.999} \\
             & \textit{Ours (def. only)} & pos+bbox\_wh & \textbf{0.058} & \textbf{0.183} & \textbf{2.286} & \textbf{0.008} & \textbf{0.025} & \textbf{0.966} & \textbf{0.996} & \textbf{0.999} \\
    
            \end{tabular}%
            }
            \caption{\textbf{Ablation results across different language embedding strategies}. All experiments in this table use the ``pos$+$bbox\_wh" positional embedding, which is a simple MLP that accepts the bounding box centre coordinates, width, and height as inputs. \textit{Ctrl. zeros} is a control experiment that replaces each object's features with zeros, \textit{def. only} uses just the synset definition for that object's label, and \textit{def $+$ sz\_rel} additionally includes a relative size clause (see section \ref{sec:object-encoder} for details).}
            \label{tab:language-ablation}
        \end{table*}
        
    \subsubsection{Language Embedding Ablation}
    \label{sec:language-ablation}
        To show the effectiveness of using language embeddings as a means of incorporating useful information to the model, we perform an ablation study across two different constructions of the phrasing passed to the language embedding module, which form the features used for the detected objects. The results are shown in table \ref{tab:language-ablation}.
        
        
        All results beat the baseline method, and the ablation shows a clear improvement in performance when using the definition of an object label only. Surprisingly, the use of a more detailed phrase containing the relative size clause is not successful in improving performance significantly compared to the control experiment. We hypothesise that this may be due to CLIP not having encountered such comparative phrases in its training dataset of image captions, and that image captions that directly relate the apparent size of objects specifically are rare. Instead, our results show that CLIP is better suited to the task of providing world knowledge about objects rather than understanding of the specific relationships between them, with the latter being better captured by the inclusion of object self-attention within our ObjCAViT block (see section \ref{sec:obj-sa-ablation}).
    
    \subsubsection{Positional Embedding Ablation}
    \label{sec:positional-embedding-ablation}
        As outlined in section \ref{sec:background}, research into monocular depth perception in the biological domain has shown a variety of different depth cues that are known to be helpful. In particular, the relative and familiar size cues, which compare the apparent sizes of objects to either a prior model of the object's true size or the apparent size of other objects in the scene, provide useful information, and have been directly leveraged for computer monocular depth estimation \cite{auty_monocular_2022}.
        
        To permit the model to make use of these cues, we alter our positional embedding model. Following \cite{sarlin_superglue_2020}, we use a simple MLP to transform 2-dimensional x/y bounding box centre coordinates to 128-dimensional positional embeddings, that get added to the object or patch embeddings. We then extend this to include apparent size information by providing the bounding box width and height to the positional encoder as a further two channels alongside the centre coordinates.
        
        \begin{table*}[]
            \centering
            \resizebox{0.925\textwidth}{!}{%
            \begin{tabular}{c|ll|ccccc|ccc}
            \multicolumn{1}{l|}{Dataset} & Lang. & Pos. emb. & $\downarrow$ Abs. Rel & $\downarrow$ Sq. Rel & $\downarrow$ RMS & $\downarrow$ RMSL & $\downarrow$ Log10 & $\delta ^1$ & $\delta ^2$ & $\delta ^3$ \\ \hline
            \multirow{5}{*}{\textbf{NYUv2}} & \multicolumn{2}{c|}{\textit{Baseline (AdaBins Reimpl.)}} & 0.123 & 0.077 & 0.428 & 0.028 & 0.052 & 0.858 & 0.978 & 0.995 \\ \cdashline{2-11} 
             & def. & pos & 0.105 & 0.059 & 0.370 & \textbf{0.021} & \textbf{0.044} & 0.900 & \textbf{0.986} & 0.996 \\
             & def. & pos+bbox\_wh & \textbf{0.103} & \textbf{0.056} & \textbf{0.368} & \textbf{0.021} & \textbf{0.044} & \textbf{0.902} & 0.985 & \textbf{0.997} \\ \cdashline{2-11} 
             & def.+sz & pos & 0.105 & \textbf{0.059} & \textbf{0.371} & \textbf{0.022} & \textbf{0.044} & 0.897 & \textbf{0.984} & 0.996 \\
             & def.+sz & pos+bbox\_wh & \textbf{0.104} & \textbf{0.059} & 0.373 & \textbf{0.022} & \textbf{0.044} & \textbf{0.898} & \textbf{0.984} & \textbf{0.997} \\ \hline
            \multirow{5}{*}{\textbf{KITTI}} & \multicolumn{2}{c|}{\textit{Baseline (AdaBins Reimpl.)}} & 0.058 & 0.192 & 2.355 & 0.009 & 0.025 & 0.965 & 0.995 & 0.999 \\ \cdashline{2-11} 
             & def. & pos & 0.060 & 0.197 & 2.443 & 0.010 & 0.026 & 0.960 & 0.994 & \textbf{0.999} \\
             & def. & pos+bbox\_wh & \textbf{0.058} & \textbf{0.183} & \textbf{2.286} & \textbf{0.008} & \textbf{0.025} & \textbf{0.966} & \textbf{0.996} & \textbf{0.999} \\ \cdashline{2-11} 
             & def.+sz & pos & \textbf{0.057} & 0.190 & 2.357 & \textbf{0.009} & \textbf{0.025} & 0.964 & \textbf{0.995} & \textbf{0.999} \\
             & def.+sz & pos+bbox\_wh & 0.058 & \textbf{0.188} & \textbf{2.304} & \textbf{0.009} & \textbf{0.025} & \textbf{0.965} & \textbf{0.995} & \textbf{0.999}
            \end{tabular}%
            }
            \caption{\textbf{Ablation study} for \textbf{different positional embedding methods}. We find that the inclusion of size information as an input to the positional embedding model (``pos+bbox\_wh") is an important factor in improving performance, in comparison to the position-only positional embeddings (``pos"). Best results of the positional embedding strategies tried for each language embedding and dataset used are highlighted in bold.}
            \label{tab:positional-ablation}
        \end{table*}
        
        The results of this ablation can be seen in table \ref{tab:positional-ablation}. We find that for most metrics, the inclusion of object size as an input to the positional embedding MLP (``pos+bbox\_wh") improves performance, in both the indoor and outdoor domains (NYUv2 and KITTI respectively). This is particularly significant when using only the definition of the object as the input to the language model, and not the relative size clause. As shown in section \ref{sec:language-ablation}, the use of the relative size clause is too complex for the language model to represent, and it confounds performance.
        
        Additionally, the addition of the bounding box dimensions has an increased effect in the KITTI benchmark. We hypothesise that this is due to the subject matter and depth range captured in KITTI: perspective, and the resultant changes in apparent size, are much more noticeable over KITTI's larger depth range of $0.001m - 80m$, compared to NYUv2's range of $0.001m-10m$.
        
    \subsubsection{Object Self-Attention ablation}
    \label{sec:obj-sa-ablation}
        A large part of our motivation is the idea that the semantic relationships between objects are relevant to depth. As discussed in section \ref{sec:language-ablation}, we hypothesise that the CLIP language model is not able to adequately capture the relationships between specific objects in a scene, leaving that task instead to the self-attention between objects.
        
        Table \ref{tab:obj-sa-ablation} shows results both with and without the use of object self-attention within the ObjCAViT module. 
        When self-attention is not used, the input features and their positional embeddings are simply forwarded to the cross-attention layer.
        
        \begin{table}[]
            \resizebox{\linewidth}{!}{%
            \begin{tabular}{lll|ccc|ccc}
                Pos. emb. & Lang. & SA? & $\downarrow$ Abs. Rel & $\downarrow$ RMS & $\downarrow$ Log10 & $\delta ^1$ & $\delta ^2$ & $\delta ^3$ \\ \hline
                \multicolumn{3}{l|}{\textit{Baseline (AdaBins Reimpl.)}} & 0.123 & 0.428 & 0.052 & 0.858 & 0.978 & 0.995 \\ \hline
                \multirow{2}{*}{pos} & \multirow{2}{*}{def.} & Y & \textbf{0.105} & 0.370 & \textbf{0.044} & \textbf{0.900} & \textbf{0.986} & 0.996 \\
                 &  & N & \textbf{0.105} & \textbf{0.368} & \textbf{0.044} & 0.899 & 0.985 & \textbf{0.997} \\ \hline
                \multirow{2}{*}{pos} & \multirow{2}{*}{def.+sz} & Y & \textbf{0.105} & 0.371 & \textbf{0.044} & 0.897 & 0.984 & \textbf{0.996} \\
                 &  & N & \textbf{0.105} & \textbf{0.367} & \textbf{0.044} & \textbf{0.900} & \textbf{0.985} & \textbf{0.996} \\ \hline
                \multirow{2}{*}{pos+bbox\_wh} & \multirow{2}{*}{def.} & Y & \textbf{0.103} & \textbf{0.368} & \textbf{0.044} & \textbf{0.902} & \textbf{0.985} & \textbf{0.997} \\
                 &  & N & 0.106 & 0.371 & 0.045 & 0.899 & \textbf{0.985} & \textbf{0.997} \\ \hline
                \multirow{2}{*}{pos+bbox\_wh} & \multirow{2}{*}{def.+sz} & Y & \textbf{0.104} & 0.373 & \textbf{0.044} & \textbf{0.898} & 0.984 & \textbf{0.997} \\
                 &  & N & 0.106 & \textbf{0.368} & \textbf{0.044} & \textbf{0.898} & \textbf{0.986} & \textbf{0.997}
            \end{tabular}%
            }
            \caption{\textbf{Object self-attention ablation} on \textbf{NYUv2} with each of the different positional embeddings and language phrasings tried, showing results both with and without object self-attention (``SA?"). Notable is that the difference is only significant for \textbf{one configuration} of model: using both position and size in computing positional embeddings, and using only definitions for phrases. We discuss this more in section \ref{sec:obj-sa-ablation}.}
            \label{tab:obj-sa-ablation}
        \end{table}
        
        Our results show that the difference with and without the use of the object self-attention is only significant when the correct language and positional model is used: using the combined location + bounding box dimension input for the positional encoder, and using only the definition (without relative size clause) for the input to the language model for a given object. As shown in section \ref{sec:language-ablation}, CLIP hurts performance if used to relate multiple objects together. When using only the position of the object bounding box as the input to the positional encoder, we observe that there is little to no difference when adding the object self-attention, (see sec. \ref{sec:positional-embedding-ablation}). This implies that the relationships modelled by the self-attention depend significantly on the apparent size as well as the position of an object in a scene, mirroring the observations of MDE in the biological domain (see section \ref{sec:background}).

    \subsubsection{Comparison to SOTA}
    \label{sec:sota-comparison}
        We provide a comparison with SOTA methods on NYUv2 in table \ref{tab:sota-nyu}. We observe that our performance matches or exceeds previous methods that use standard convolutional backbones, including our re-implementation of AdaBins that uses the same batch size as our method. We also note that our $\sim 30M$ param EfficientNet-B5 \cite{tan_efficientnet_2019} backbone is considerably smaller than the $197M$ to $220M$ or more params of the BinsFormer \cite{li_binsformer_2022}, PixelFormer \cite{agarwal_attention_2022} and DepthFormer \cite{li_depthformer_2022} backbones, and that our method is more modular and may therefore be easily applied to any dense feature extractor.
        
        \begin{table}[]
            \resizebox{\linewidth}{!}{%
            \begin{tabular}{l|ccc|ccc}
                Model & $\downarrow$ Abs. Rel & $\downarrow$ RMS & $\downarrow$ Log10 & $\delta ^1$ & $\delta ^2$ & $\delta ^3$ \\ \hline
                Eigen et al. \cite{eigen_depth_2014} & 0.158 & 0.641 & - & 0.769 & 0.950 & 0.988 \\
                Laina et al. \cite{laina_deeper_2016} & 0.127 & 0.573 & 0.055 & 0.811 & 0.953 & 0.988 \\
                DORN \cite{fu_deep_2018} & 0.115 & 0.509 & 0.051 & 0.828 & 0.965 & 0.992 \\
                BTS \cite{lee_big_2019} & 0.110 & 0.392 & 0.047 & 0.885 & 0.978 & 0.994 \\
                AdaBins (reported, bs=16) & 0.103 & 0.364 & 0.044 & 0.903 & 0.984 & {\ul 0.997} \\
                AdaBins (reimpl., bs=8) & 0.123 & 0.428 & 0.052 & 0.858 & 0.978 & 0.995 \\
                DepthFormer \cite{li_depthformer_2022} & 0.096 & 0.339 & 0.041 & 0.921 & {\ul 0.989} & \textbf{0.998} \\
                BinsFormer (Sw-L) \cite{li_binsformer_2022} & {\ul 0.094} & {\ul 0.330} & {\ul 0.040} & {\ul 0.925} & {\ul 0.989} & {\ul 0.997} \\
                PixelFormer \cite{agarwal_attention_2022} & \textbf{0.090} & \textbf{0.322} & \textbf{0.039} & \textbf{0.929} & \textbf{0.991} & \textbf{0.998} \\ \hline
                \textbf{Ours (def, pos+bbox\_wh)} & 0.103 & 0.368 & 0.044 & 0.902 & 0.985 & {\ul 0.997}
                \end{tabular}%
            }
            \caption{Comparison of our method to SOTA methods from the literature on NYUv2, best and 2nd-best results in bold and underlined respectively. We note that our performance exceeds our reimplementation of AdaBins that was retrained with the same batch size as our method, and gives competitive results with other methods in the literature that use considerably larger backbones than ours. Numbers are sourced from the respective source papers.}
            \label{tab:sota-nyu}
        \end{table}
    

%% file: sections/conclusion.tex
\section{Conclusion \& Future Work}
\label{sec:conclusion}
In this work, we have introduced ObjCAViT, a novel method for improving monocular depth estimation by using auxiliary information provided by frozen off-the-shelf object detection and language models. We compel the network to pay attention to objects in a scene, sourcing assumptions about the nature of the objects from general-purpose language models and then using self-attention to learn to model the relationships of those objects to one another. Our ObjCAViT block permits the model to relate the object and inter-object semantics to the visual features in the image, and we achieve excellent results that exceed our baseline and are competitive with models much larger than ours. We also find that incorporating the apparent size of objects to their feature representation contributes significantly to the improvement. Our ablation studies confirm that each component of our final model contributes to the performance, though we acknowledge the limitations of enforcing a specific bias on the model and of relying on the accuracy of an off-the-shelf model to provide object information that is assumed to be reliable.

Our future work will aim to alter the architecture of the backbone used to extract the dense image features. Very recent work \cite{agarwal_attention_2022} has begun to use all-transformer architectures for visual feature extraction; we expect that combining object and visual features at multiple stages of the image encoding process will allow the extraction of more depth-relevant image features, which will in turn lead to further increases in performance. While the CLIP model used in our work is general-purpose, we hypothesise that the use of language models trained for reasoning-like tasks will further improve performance.

%% file: sections/app-positional-encoding-arch.tex
\section{Positional Encoder Architecture}
\label{app:positional-architecture}
    The positional encoders used are simple MLPs with LeakyReLU activations. The position-only positional encoder (``pos") is adapted from the positional encoder used in SuperGlue \cite{sarlin_superglue_2020}, and has the following structure (using PyTorch notation):
    \begin{verbatim}
        nn.Sequential(
            nn.Linear(2, 32, bias=True),
            nn.LeakyReLU(),
            nn.Linear(32, 64, bias=True),
            nn.LeakyReLU(),
            nn.Linear(64, 128, bias=True),
            nn.LeakyReLU(),
            nn.Linear(128, 256, bias=True),
            nn.LeakyReLU(),
            nn.Linear(256, 128, bias=True)
        )
    \end{verbatim}
    
    The joint position-and-apparent size positional encoder (``pos+bbox\_wh") additionally accepts the width and height, in pixels, of the object bounding boxes. It has the following structure:
    \begin{verbatim}
        nn.Sequential(
            nn.Linear(4, 32, bias=True),
            nn.LeakyReLU(),
            nn.Linear(32, 64, bias=True),
            nn.LeakyReLU(),
            nn.Linear(64, 128, bias=True),
            nn.LeakyReLU(),
            nn.Linear(128, 256, bias=True),
            nn.LeakyReLU(),
            nn.Linear(256, 128, bias=True)
        )
    \end{verbatim}

%% file: sections/app-additional-examples.tex
\section{Additional Qualitative Examples}
\begin{figure*}
        \centering
        \includegraphics[width=\textwidth]{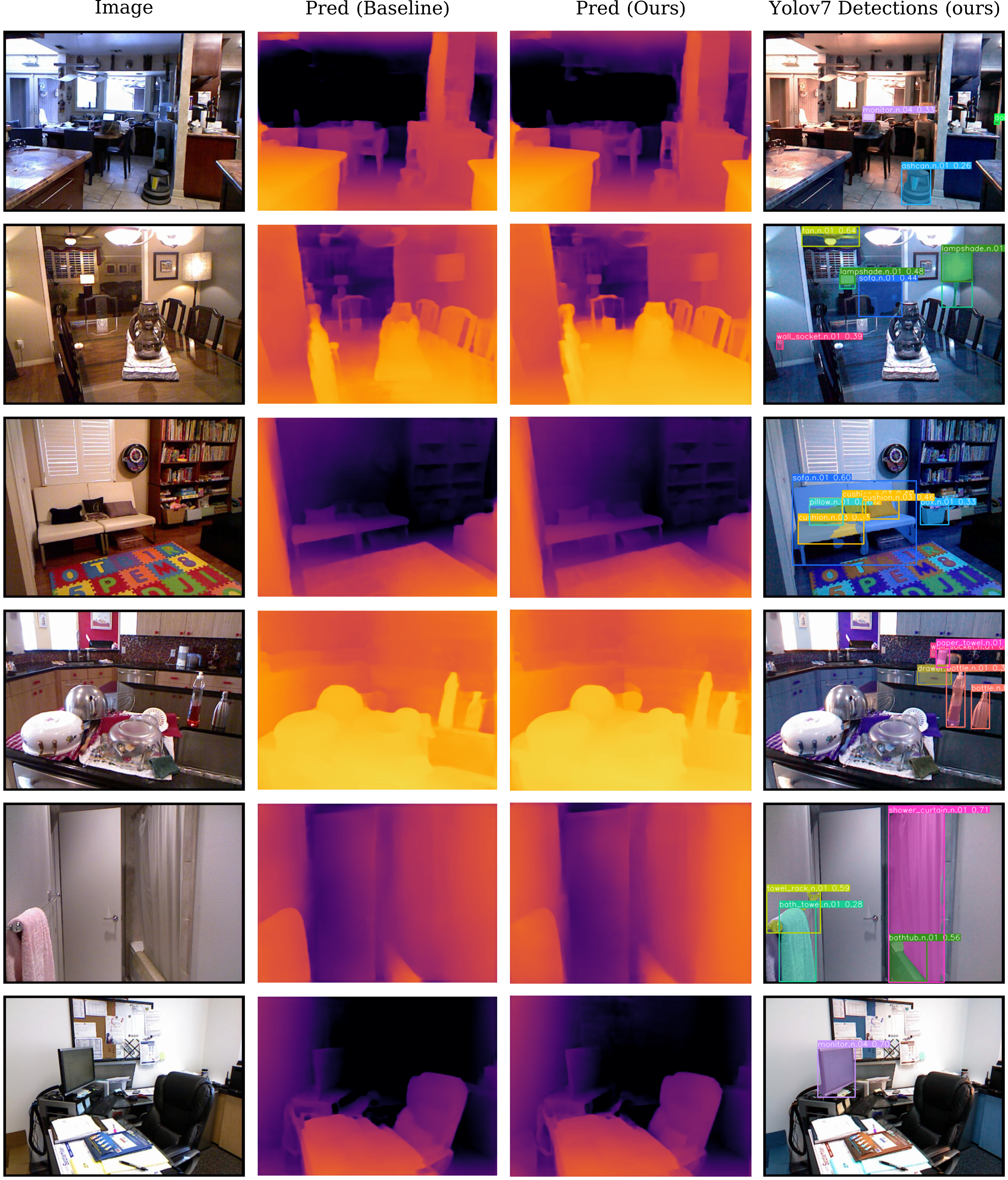}
        \caption{Qualitative results from running inference on the NYUv2 test set. Test-time augmentation is not used here, but is used for computations of the metrics reported in the paper (as is standard in the existing literature \cite{bhat_adabins_2020, lee_big_2019}). ``Pred (Baseline)" results were produced by the reimplemented AdaBins (denoted as ``AdaBins (reimpl.)" in the main paper. ``Pred (Ours)" results were produced by our best model, which used a positional encoder with both position and object bounding box dimensions provided as input (\textit{pos+bbox wh}) and used only the WordNet definition as the language input for detected objects (\textit{def.}).}
        \label{fig:nyu_qualitative_1}
    \end{figure*}

\begin{figure*}
        \centering
        \includegraphics[width=\textwidth]{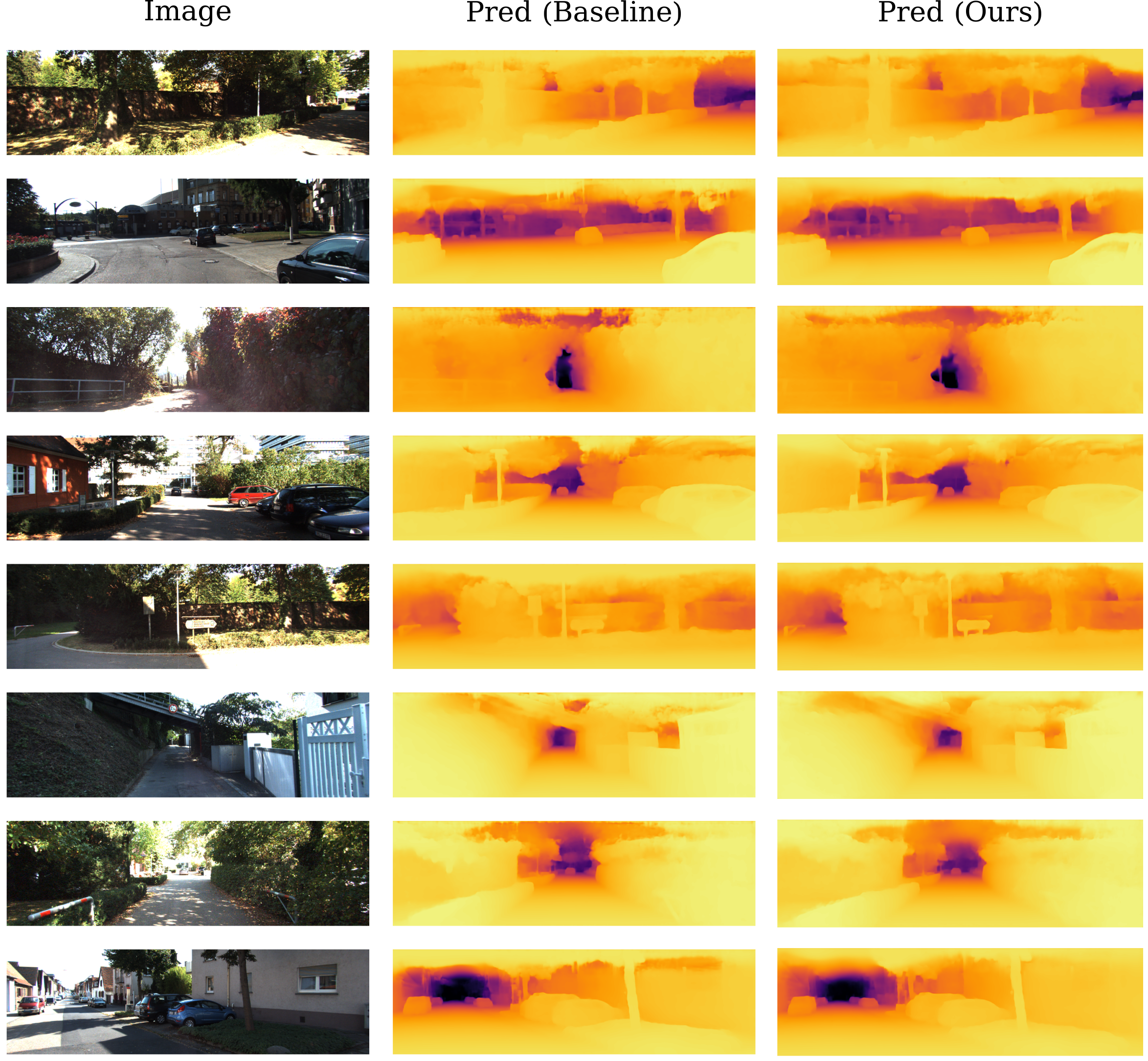}
        \caption{Qualitative results from running inference on the KITTI test set. Object detections are not shown to allow predictions to be shown larger. Inference settings and models used are as described in the caption of figure \ref{fig:nyu_qualitative_1}.}
        \label{fig:kitti_qualitative_1}
    \end{figure*}

    This section includes some additional qualitative examples that were not included in the main paper for lack of space. Figure \ref{fig:nyu_qualitative_1} contains examples from the NYUv2 test split, and figure \ref{fig:kitti_qualitative_1} contains examples from the KITTI test split.
    
